\documentclass{article}

\usepackage{arxiv}

\usepackage[utf8]{inputenc} 
\usepackage[T1]{fontenc}    
\usepackage{hyperref}       
\usepackage{url}            
\usepackage{booktabs}       
\usepackage{amsfonts}       
\usepackage{nicefrac}       
\usepackage{microtype}      
\usepackage{lipsum}

\usepackage{graphicx}
\usepackage{amssymb}
\usepackage{amsmath}
\usepackage{multirow}

\usepackage[ruled,vlined]{algorithm2e}
\usepackage{algorithmic}

\title{BraidNet: procedural generation of neural networks for image classification problems using braid theory}

\author{
  Olga~Lukyanova\thanks{Corresponding author} \\
  Computing Center \\
  Far Eastern Branch\\ 
  Russian Academy of Sciences\\
  680000, Khabarovsk, Russia\\
  \texttt{ollukyan@gmail.com} \\
   \And
  Oleg~Nikitin\\
  Computing Center \\
  Far Eastern Branch\\ 
  Russian Academy of Sciences\\
  680000, Khabarovsk, Russia\\
  \texttt{olegioner@gmail.com} \\
  \And
 Alex~Kunin \\
  Computing Center \\
  Far Eastern Branch\\ 
  Russian Academy of Sciences\\
  680000, Khabarovsk, Russia\\
  \texttt{alexkunin88@gmail.com} \\
}

\begin{document}
\maketitle

\begin{abstract}
In this article, we propose the approach to procedural optimization of a neural network, based on the combination of information theory and braid theory. The network studied in the article implemented with the intersections between the braid strands, as well as simplified networks (a network with strands without intersections and a simple convolutional deep neural network), are used to solve various problems of multiclass image classification that allow us to analyze the comparative effectiveness of the proposed architecture. The simulation results showed BraidNet's comparative advantage in learning speed and classification accuracy.
\end{abstract}





\section{Introduction} The architecture of neural networks is selected by studying their accuracy and ability of generalization. This approach is not optimal and requires a lot of time and computational resources. So, it can be useful to replace it by the automatic optimization of neural network architectures. Automatic approaches imply the use of algorithmic (procedural) methods for the generation of neural networks, that is, the application of rules and procedures that create certain \cite{1} sequences. This approach will be useful for generating deep neural network architectures, since the optimal setting of their structure is nontrivial and directly depends on the problem being solved. Algorithmic search of neural network architectures
is an actively developing research topic of current interest \cite{2}.

In this article, the authors propose the use of modern mathematical methods to optimize the structure of neural networks. It also provides an overview of the application of braid theory and information theory to improve the performance of neural networks. In previous research \cite{5} procedural generation techniques based on matrix filters and braid theory were studied by authors. In the present article, we modify such an approach to generate and optimize BraidNets.
\section{Procedural generation of neural networks} Usually, experts select the structures and hyperparameters of deep neural networks manually. This can lead to highly efficient specialized architectures. The disadvantage of this method is the need for long-lasting experiments to solve each specific problem. To overcome this, the Neural Architecture Search (NAS) methods have been proposed, such as NAS-RL \cite{7} and MetaQNN \cite{8}, which use reinforcement learning, a method of machine learning, in which a model that does not have information about the system but has the ability to perform any actions in it to optimize a parameter is trained. NAS-RL and MetaQNN require a lot of computation in the selection process to assess the effectiveness of the resulting architecture. These algorithms imply the direct setting of the network modules' structure in the optimization process.

NAS approaches usually involve direct specification of optimized architectures, which often expands the search space and complicates the optimization. Procedural generation of architectures complements these approaches and involves the creation of architectures based on algorithmic rules. Such approaches in evolutionary optimization are called indirect coding -- the generation rule is encoded, not the direct result. Examples of such method are: Convolutional Neural fabrics (CNF) \cite{10}, PathNet \cite{11}, and Budgeted Super Networks (BSN) \cite{12}. All these approaches are united by the method of forming architectures, which implies the generation of types of modules and connections between them based on algorithmic approaches.

In general, all of the compared approaches require complex computations to optimize the network topology, and they do not support multi-column and parallel architectures. Many of them involve one or the other of the brute force methods. To build optimal architectures, there are not enough analytical estimates of information indicators of network performance and, nowadays, the selection is made empirically. This makes the problems of procedural generation of neural architectures computationally complex or suboptimal, therefore, research in the field of neural network topologies and the search for theoretical estimates of information dynamics in neural networks is an important task.

To investigate such a procedure, it is proposed to use a combination of low-dimensional topology and information theory. The use of braid theory to describe information structures and transformations, due to the ability of braids to describe information transfer paths on computational graphs, is designed to overcome the disadvantages of previous approaches since it eliminates the need for brute force.

\section{Braid theory} 
\begin{figure}
\centering 
\includegraphics[width=50mm]{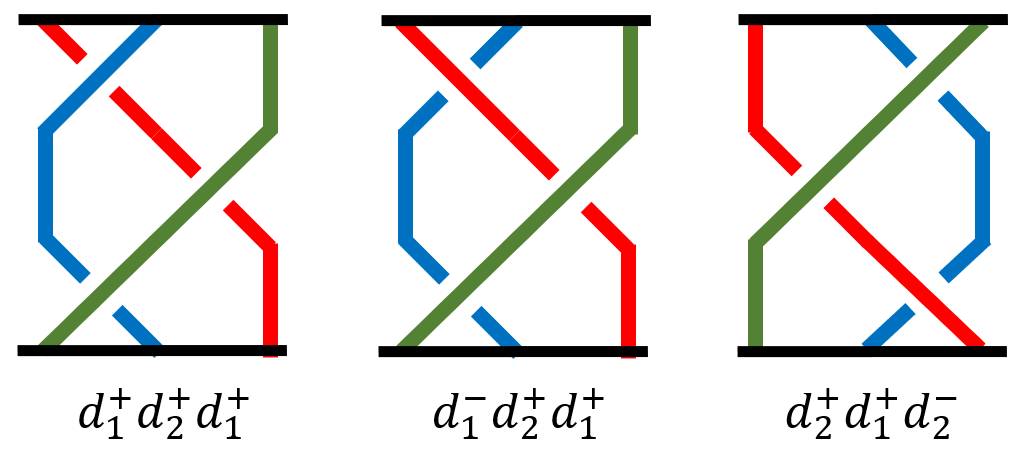}
\caption{The example of three braids and their letter descriptions.} 
\label{img:2}  
\end{figure}

Braid theory is a section of topology and algebra that studies braids (pairwise intersections of strands) and braid groups composed of their equivalence classes. A mathematical braid consists of $m$ strands (that is, curves in space) that start at $m$ points on a horizontal line and end at $m$ points on another horizontal line below \cite{25}. Examples of three braids are shown in Figure~\ref{img:2}.

\begin{figure}
\centering 
\includegraphics[width=50mm]{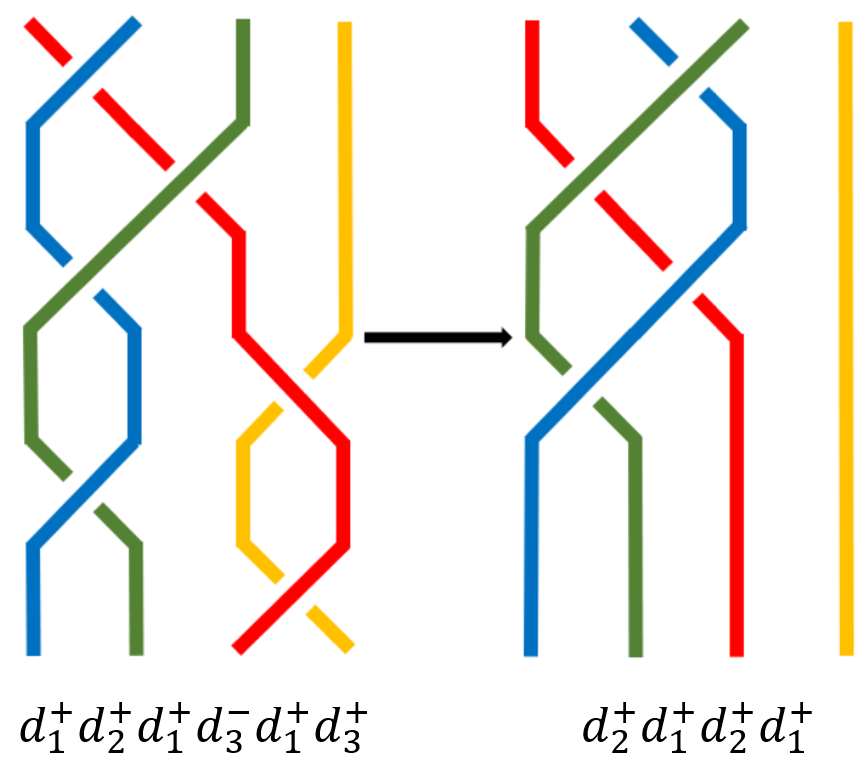}
\caption{The braid isotopy example.} 
\label{img:4}  
\end{figure}

The braid can be represented as a "braid word", i.e. the product of elements  $d_i^{\pm}$, where $d_i$ are the braid diagrams on three strands. Thus, the elements $d_i,...,d_{n-1}$ generate the group $D_n$. Let us denote by $d_i$ a braid of $m$ strands, the $i$-th strand of which passes "under" $(i+1)$-th, and the other strands have no crossings. This is a sequence of notation $d_1^+$, $d_1^-$, $d_2^+$, $d_2^-$ and so on. It can be written which strand of the braid goes under or above the other strand. A braid diagram with $m$ strands has positions from 1 to $m$. If the strand $m$ is crossed from above by the strand $m+1$, it is written as $d_m^-$. And vice versa: $d_m^+$ for the case when the strand $m+1$ passes under the strand $m$. Thus, you can "move" the strands of the braid, but you cannot disconnect them from the points at which they begin and end, cut and glue them. Various versions of braids that arise after these movements are equivalent (isotopic) to the original braid \cite{27}. So, figure~\ref{img:4} shows an example of "moving" a braid and the braid after these transformations.

In this paper, the Dehornoy's handle reduction algorithm \cite{28,31} is used to find the isotopes of the braids used in the BraidNet architecture. Interconnection structures between neural network modules, defined using braid theory, can be used to build network topologies with multiple threads (strands).

\section{Information theory} The architecture of neural networks, described using braid theory, allows pairwise comparison of the significance of diverse layers at different levels of the network using various estimates, including information theory. Below we use information entropy as an estimate of statistical significance.

\begin{figure}
\centering
\includegraphics[width=80mm]{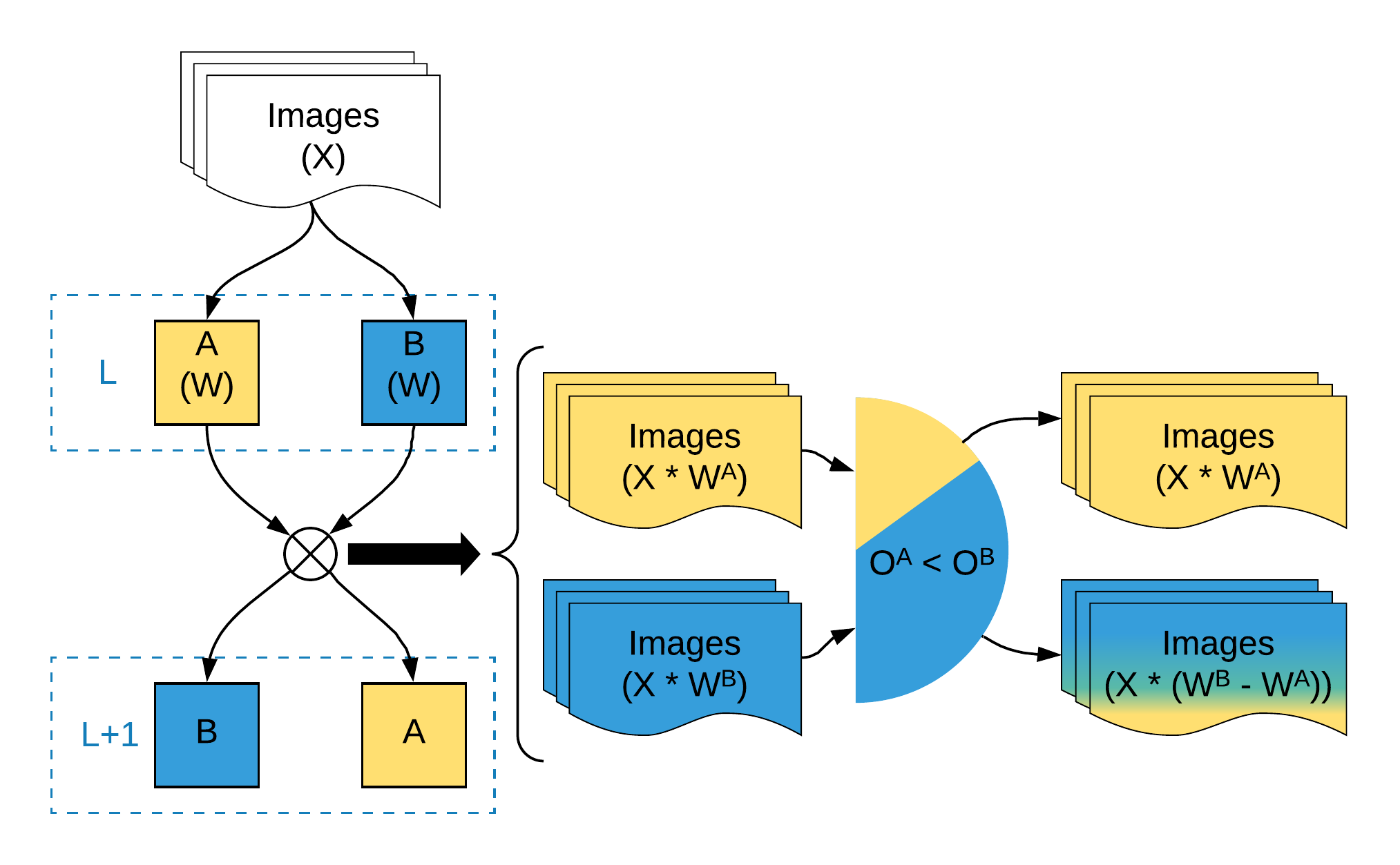}
\caption{An example of transferring information between network layers} 
\label{img:5}  
\end{figure}

\textit{Information entropy} is a measure of the uncertainty of the system that evaluates the expected (average) amount of information transmitted by determining the results of a random test \cite{35}. In this work, the properties of information entropy were applied to determine the amount of information fed on each layer of the neural network, and to select paths containing more information about the learning object. In accordance with the proposed algorithm, the active modules of the neural network and the connections between them change during the learning procedure. Intersections between active modules are specified using braid theory. The transmission structure between the layers at the intersection points is selected based on a comparison of the information entropy of active modules: the lower the information entropy, the higher the module's confidence in the classification of the data received into it. The scheme of information transfer between the network layers at the strand crossings for such a case will correspond to the figure~\ref{img:5}.

Since strands with modules $A$ and $B$ solve the problem of finding their classes among all classes $N$, then in cases when the $A$ module with high probability ("certainty") determines the information $x$ as related to its class, this information is less likely to be determined by module $B$ as belonging to its class. Subtraction of the weights of different modules is used to solve this problem. In this case, the weights change only for the module with the highest information entropy.

Thus, the information of two strands on a layer after they cross is calculated as follows:

\begin{equation}
\label{eq:eq3}
  X_{L+1} = 
  \begin{cases}
   X_L (W_L^B - W_L^A), &\text{if $O^A(X_L)<O^B(X_L)$}\\
   X_L (W_L^A - W_L^B), &\text{if $O^A(X_L)\geq O^B(X_L)$}
  \end{cases},
\end{equation}
where  \\
$A$,$B$ -- the modules, $A,B\in N$, \\
$N$ -- the number of strands (number of classes), \\
$W_L^A$, $W_L^B$ -- the weights of the modules $A$ and $B$ on the layer $L$,\\
$x\in X_L$ -- input information on layer $L$. \\

$O^A(X_L)$ defines the number of cases when the module $A$ exceeds the module $B$ in information entropy when classifying the information $x$.

\begin{equation}
  \label{eq:eq4}
  O^A(X_L) = \sum \limits_{x=1}^{X_L} max(H^A(x), H^B(x)).
\end{equation}

Information entropy of $A$ module:

\begin{equation}
  \label{eq:eq5}
  H^A(x) = p_1^A(x) \log_{2} \frac{1}{p_1^A(x)} + p_2^A(x) \log_{2} \frac{1}{p_1^A(x)},
\end{equation}
where \\
$p_1^A(x)$ -- the probability that $x$ belongs to the required class $A$, \\
$p_2^A(x)$ -- the probability that $x$ does not belong to the required class $A$. \\

Similar calculations are performed for the $B$ module according to the equations~\ref{eq:eq4}--\ref{eq:eq5}.

Thus, the calculation of the information entropy of modules in the process of training a neural network allows you to dynamically change the paths of information transfer and select the most informative modules. This kind of comparison is possible between modules only in pairs, which well complements the pairwise character of intersections in braid structures described in the previous chapter of this work.

\section{BraidNet deep neural network architecture} The structures of neural networks set by the braids allow the calculation and comparison of information entropy of weights in network modules. Thus, a combination of braid theory and information theory can be applied to automatically form and dynamically change the structure of the neural network. This allows not only to select the most successfully trained network modules, but also to do it automatically, avoiding manual iteration over the sequences of comparisons between modules. At the same time, based on research in \cite{2}, it can be assumed that changing the network topology within one isotopy leads to similar learning results and contributes to information consistency within the layer permutations. To test these assumptions, the BraidNet learning algorithm described below has been developed.

\begin{figure}
\centering
\includegraphics[width=\textwidth]{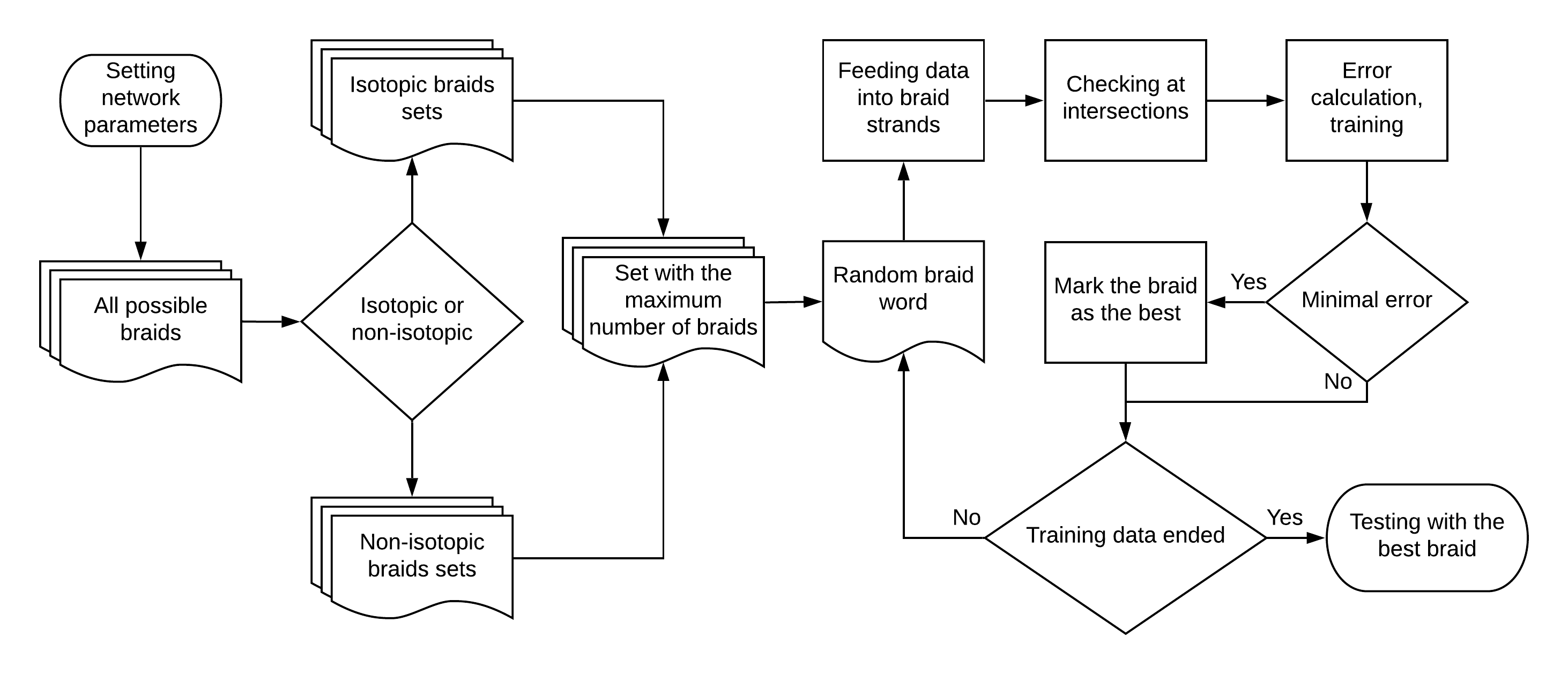}
\caption{A general algorithm for the functioning of the BraidNet architecture}
\label{img:6}
\end{figure}

The general algorithm for the functioning of the BraidNet architecture assumes the execution of the stages shown in Figure~\ref{img:6}.

The network parameters are set by the number of layers and the number of strands of the braid, where each of the strands solves the problem of finding its own data class among all input examples ("one against all" problem). The number of strands corresponds to the number of classes. Further, all possible existing options for braids are determined for the given parameters, that is, the number of strands and their length. Then a set of braids is formed: in the BraidNet architecture with isotopic braids, for each of the obtained braids, a set of braids isotopic to it is found; in the case of non-isotopic braids, for each of the resulting braids, a set of non-isotopic braids is formed in such a way that each of the braids in the set is non-isotopic to any other braid from this set. After determining the set of braids, a set is selected with the maximum number of variants of neural network structures in it. If there are several such sets, then the set among them is selected at random. Next, the structure of the neural network (braid words) is randomly selected from the resulting set.

The input data is fed to each of the braid strands at the same time. After that, a check is carried out for the presence of intersections: if they exist, at the intersections the information entropy of the modules whose strands participate in the intersection is calculated, and the weights of the module with less information are subtracted from the module with the highest information entropy. The data of these modules is transmitted further over the network, and then, the residual of the prediction from the fact is calculated separately for each of the strands and the training is performed using the backpropagation method. The procedure is repeated for all training data, starting from step 5 ("Random braid word"). After that, testing is performed with the braid, the results for which had the smallest error after the training.

\begin{figure}
\centering
\includegraphics[width=60mm]{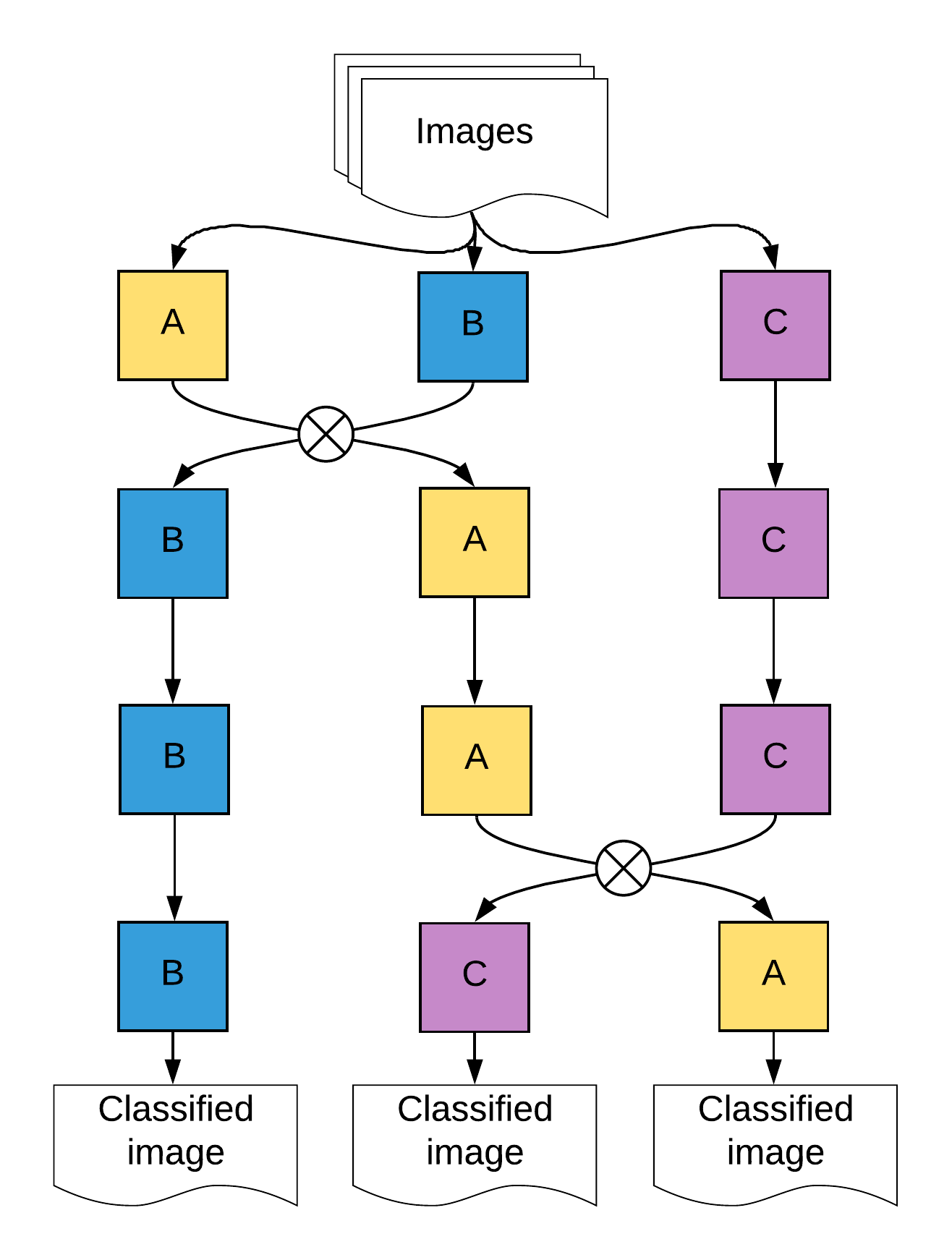}
\caption{An example of a neural network architecture BraidNet, based on the braid theory, consisting of three braid strands}
\label{img:7}
\end{figure}

An example of the resulting architecture of the deep neural network is shown in Figure~\ref{img:7}. This network consists of braid strands, the number of which is determined by the number of classes recognized by the network. Each strand learns to recognize the class assigned to it.

The combination of structures defined by braids and dynamic modification of the computation graph, using information theory, makes it possible to obtain structures that naturally allow solving problems of multiclass classification, since the presence of many threads in the network helps learning individual strands for each class, and mixing and evaluating information -- to combine some common features for different classes.

To solve classification problems in this work, the BraidNet architecture was used, each strand of which consisted of four modules. The first two modules contained convolutional layers with a kernel equal to 5, and activation layers and pooling with a step equal to 2. The next two modules consisted of fully connected layers. At the end of each strand, a logistic activation function was applied.

\section{Simulation results} To assess the performance of the BraidNet network and compare it with other architectures in the context of multiclass classification problems, the neural networks were applied to classical image recognition problems -- MNIST, SVHN, CIFAR10. The BraidNet network was implemented with crossings between strands specified by the braid. For comparison, the neural network architecture 3DNN was chosen -- a deep neural network of three parallel threads, discussed in \cite{5}, and CNN is a simple convolutional deep neural network. All parameters on the layers of the compared networks were the same. The architecture of the deep neural network BraidNet included the construction of a network connection graph based on the braid theory and it's isotopic properties, as well as dynamic optimization of the network structure based on information theory. Two variants of the BraidNet architecture were investigated - with and without braid isotopy.

To define the active network modules, binary matrix filters were used -- algorithms that use a certain matrix, highlighting (or removing) from the original object some part with the given properties \cite{5}. Such filters define the active modules of the network, thereby changing the path of information transfer between layers. They are binary matrices, where "1" means the activation of the module located in the given corresponding layer, and "0" means the exclusion of the module from the network. The resulting binary matrix is called the filter of active modules $\it F_a$. The final architecture of the active modules of the neural network is represented by the matrix of active modules $\it M_{act}$. It is obtained using the Hadamard product of the binary filter $\it F_a$ and the matrix of possible active modules $\it M_{full}$ of the neural network:
          
\begin{equation}
\label{eq:eq6}
M_{act} = M_{full} \circ F_a
\end{equation}

The MNIST, SVHN and CIFAR10 tasks involve the recognition of different kinds of image classes. The MNIST problem recognizes handwritten numbers, the SVHN recognizes the numbers on the house number indicators, in CIFAR10, images belong to 10 different classes of objects, such as an airplane, a cat, a dog, and a truck. Of all these tasks, it is CIFAR10 that is the most difficult, since it contains very heterogeneous information.

\begin{figure}
\centering
\includegraphics[width=\textwidth]{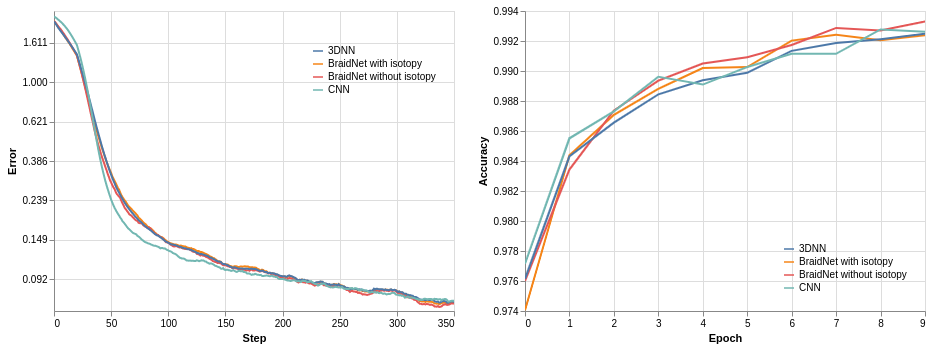}
\caption{The results of modeling the MNIST problem}
\label{img:9}
\end{figure}

According to the results of training networks on the MNIST problem, as on the simplest problem, it can be seen that all networks showed approximately the same efficiency and learning rate, but BraidNet took slightly fewer training epochs. At the same time, the results of the dynamics of the loss function on the test sample (Figure~\ref{img:9}) indicate a somewhat better ability of BraidNet to generalize and less tendency to overfit.

\begin{figure}
\centering
\includegraphics[width=\textwidth]{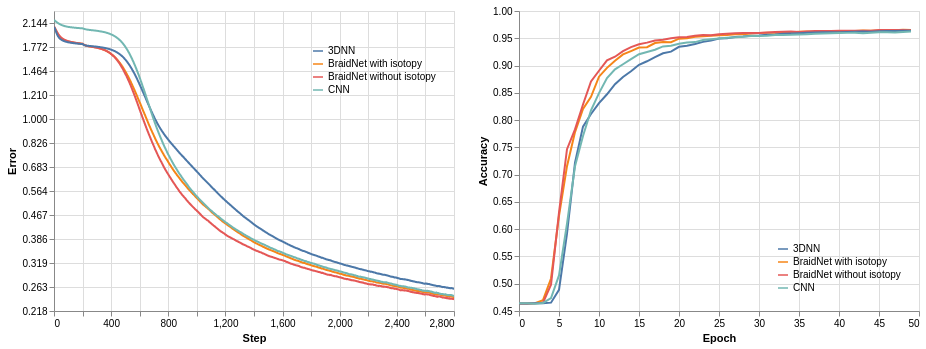}
\caption{The results of modeling the SVHN problem}
\label{img:10}
\end{figure}

The learning results for a slightly more complex SVHN numbers recognition problem are shown in Figure~\ref{img:10}. In this case, we can observe significantly different dynamics in the training of BraidNet and deep convolutional networks. The BraidNet architecture allows you to quickly achieve an increase in accuracy on a test sample, due to the presence of dynamics in the selection of modules in the learning process (Table~\ref{tab:2}).

\begin{table}
\caption{Network training results, epoch number}
\centering 
\label{tab:2}
\begin{tabular}{|c|c|c|c|}
\hline
Architecture & MNIST, 99\% & SVHN 96\% & CIFAR10 88\% \\
\hline
CNN & 6 & 43 & 94 \\
3DNN & 7 & 37 & 90 \\
BraidNet, with isotopies & \textbf{5} & \textbf{32} & 85 \\
BraidNet, without isotopies & \textbf{5} & \textbf{32} & \textbf{80} \\
\hline
\end{tabular}
\end{table}

The table shows that networks based on the BraidNet architecture are able to learn faster than convolutional neural networks. However, comparing BraidNet with isotopic and non-isotopic braids showed the same results in the MNIST and SVHN problems, as well as the best results of training the network with non-isotopic braids in the CIFAR10 problem. Thus, we can conclude that isotopy preserves information homogeneity and leads to overfitting.

The most challenging task studied, CIFAR10, compared the efficiency of isotopy among the studied braid sets. Based on the results presented in table~\ref{tab:2}, it can be concluded that the maintenance of isotopy among the network topologies contributes to the maintenance of information homogeneity in the learning process, which slows down the search for optimal values. Braids without isotopies help to bypass local minima in the learning process.

\begin{figure}
\centering
\includegraphics[width=\textwidth]{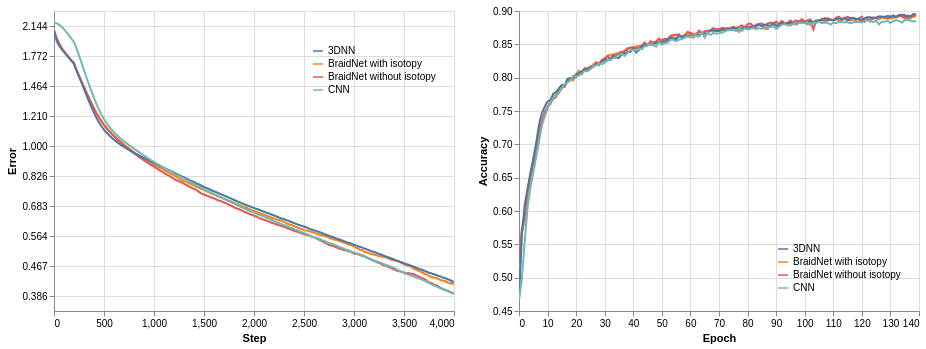}
\caption{The results of modeling the CIFAR10 problem}
\label{img:11}
\end{figure}

Figure~\ref{img:11} shows that, in general, the results of training all networks on the CIFAR10 problem are very close, but the network with the BraidNet architecture shows a more dynamic decrease in the values of the error function during the learning process on the training and test sets.

Comparing BraidNet with other procedurally generated topology networks shows comparable results (Table~\ref{tab:3}).

\begin{table}
\caption{Comparative analysis of the operation of the BraidNet network with other architectures using the example of the CIFAR10 task (Error (CIFAR10, \%) -- percentage of the network training error on the CIFAR10 task, GPU days -- the number of days spent on training the model on the GPU)}
\centering \small\label{tab:3}
\begin{tabular}{|l|c|c|c|}
\hline
Approach & Number of parameters, million & Error (CIFAR10,\%) & GPU days \\
\hline
ResNet & 1.7 & 6.43 & 0.65 \\
NAS-RL & 37.4 & 3.65 & 22400 \\
MetaQNN & 11.18 & 6.92 & 100 \\
Large-scale Evolution & 5.4 & 5.4 & 2600 \\
AmoebaNet & 3.2 & 3.34 & 3150 \\
ShakeDrop & 26.2 & 2.67 & - \\
NASBOT & 8.69 & 12.3 & 1.7 \\
DARTS & 3.3 & 2.76 & 4 \\
SNAS & 2.9 & 2.98 & 1.5 \\
PathNet & 13 & 60.2 & - \\
PyramidNet & 26 & 3.31 & 84 \\
Budgeted SuperNets & 7.56 & 5.12 & - \\
CNF & 16 & 7.43 & 273 \\
FractalNet & 38.6 & 7.27 & - \\
EnvelopeNet & 10 & 3.33 & 0.25 \\
BraidNet & 6.5 & 14.3 & 0.05 \\
\hline
\end{tabular}
\end{table}

In the MNIST problem, the lag in terms of final accuracy was 0.58\%, in the SVHN problem -- 2.82\%, in the CIFAR10 problem -- 7.03\%. At the same time, it should be noted that the BraidNet approach requires fewer computing resources and can be performed even on a home computer with a modern GPU. The BraidNet network contains much fewer parameters and also does not imply recalculation of all these parameters during training. In fact, for a task with ten classes, no more than half of the 6.5 million parameters are trained at the same time.

Thus, the simulation results showed the comparative advantage of BraidNet in learning speed and classification accuracy. Isotopic properties in this neural network architecture require further study. The BraidNet architecture is a promising algorithm for further application and research in complex problems of pattern recognition, including the use of neuroevolutionary approaches.

\section{Conclusion} As part of research on the automatic generation of neural network structures, the authors proposed and studied the BraidNet approach, which combines the generation of network topology using braid theory and its optimization, using estimates from information theory. The work also implemented a number of simplified architectures to analyze the performance of the BraidNet architecture. The neural networks, such as BraidNet, 3DNN, consisted of threads without intersections, as well as a simple convolutional deep neural network (CNN) were applied to solve the multiclass image classification problems -- MNIST, SVHN, and CIFAR10. The simulation results showed the comparative advantage of BraidNet in learning speed and classification accuracy. Comparing BraidNet with other procedurally generated topology networks shows slightly lagging results with much fewer network parameters. In general, BraidNet training is less computationally intensive than similar approaches.

It should be noted that the study did not carry out additional optimization of the BraidNet network hyperparameters, such as the learning rate and dropout rate, and did not conduct experiments to introduce noise into the training set. These actions, as well as the inclusion of various layers in the network, except for convolutional ones (pooling, dropout, ReLu, etc.), can increase the final accuracy of the network. Despite the existing opportunities for improving the BraidNet network, it can be applied in practice in the form obtained during the implementation. Thus, the BraidNet approach can be used to optimize neural networks when solving classification problems.

\section*{Acknowledgement}
The computing resources of the Shared Facility Center "Data Center of FEB RAS" (Khabarovsk) were used to carry out calculations.

\bibliographystyle{unsrt}
\bibliography{mybibliography}

\end{document}